\documentclass[preprint,12pt]{elsarticle}

\usepackage{amssymb}

\usepackage{url}
\usepackage{breakurl}
\usepackage[breaklinks]{hyperref}
\usepackage{multirow}

\begin{document}
\emergencystretch 3em

\begin{frontmatter}



\title{Mixture of Self-Supervised Learning}


\author[inst1]{Aristo Renaldo Ruslim}

\affiliation[inst1]{organization={Informatics Department, Faculty of Computer Science, Brawijaya University},
            city={Malang},
            postcode={65145}, 
            state={East Java},
            country={Indonesia}}

\author[inst1]{Novanto Yudistira}
\author[inst1]{Budi Darma Setiawan}


\begin{abstract}
 Self-supervised learning is popular method because of its ability to learn features in images without using its labels and is able to overcome limited labeled datasets used in supervised learning. Self-supervised learning works by using a pretext task which will be trained on the model before being applied to a specific task. There are some examples of pretext tasks used in self-supervised learning in the field of image recognition, namely rotation prediction, solving jigsaw puzzles, and predicting relative positions on image. Previous studies have only used one type of transformation as a pretext task. This raises the question of how it affects if more than one pretext task is used and to use a gating network to combine all pretext tasks. Therefore, we propose the Gated Self-Supervised Learning method to improve image classification which use more than one transformation as pretext task and uses the Mixture of Expert architecture as a gating network in combining each pretext task so that the model automatically can study and focus more on the most useful augmentations for classification. We test performance of the proposed method in several scenarios, namely CIFAR imbalance dataset classification, adversarial perturbations, Tiny-Imagenet dataset classification, and semi-supervised learning. Moreover, there are Grad-CAM and T-SNE analysis that are used to see the proposed method for identifying important features that influence image classification and representing data for each class and separating different classes properly. Our code is in \href{https://github.com/aristorenaldo/G-SSL}{https://github.com/aristorenaldo/G-SSL}
\end{abstract}




\end{frontmatter}


\section{Introduction}
\label{sec:introduction}

Deep learning methods \cite{lecun_deep_2015} have demonstrated an excellent performance on various machine learning tasks, such as computer vision for example, image classification \cite{he_deep_2016, huang_densely_2017} and semantic segmentation \cite{long_fully_2015, sun_fully_2018}, natural language processing for example, sentiment analysis \cite{zhang_deep_2018}, and the pretrain language model \cite{devlin_bert_2019}. In general, supervised learning works by training models on large, randomly divided labeled datasets for training, validation, and testing. However, supervised learning has problems, namely depending on labeled datasets whose labeling is still done manually, generalization errors occur, and adversarial attacks \cite{liu_self-supervised_2023}. The self-supervised learning method is a very promising method for solving the problem of limited and rare labeled datasets in traditional deep learning algorithms \cite{hendrycks_using_2019}. Therefore, the self-supervised learning method is attracting attention as an alternative because of its data efficiency and generalization capabilities as well as a lot of research and development on this method \cite{liu_self-supervised_2023}.

Self-supervised learning is widely applied and shows very significant results in feature learning, especially in images \cite{chen_exploring_2021, he_momentum_2020}. Self-supervised learning works by creating additional tasks or pretext tasks \cite{jing_self-supervised_2021} which will be trained on a model before being applied to more specific tasks. In image recognition, pretext tasks \cite{jing_self-supervised_2021} are usually performed for automatic labeling to create pseudo labels as well as augmenting and transforming datasets based on the data structure and type of pretext task used then predicting pseudo labels such as rotation prediction images \cite{gidaris_unsupervised_2018}, solving jigsaw puzzles \cite{noroozi_unsupervised_2016}, and predicting the relative positions of image sections \cite{doersch_unsupervised_2015}.

The use of data augmentation techniques in image recognition can improve the model's ability to recognize and generalize unique features in images so as to prevent over-fitting of the model \cite{shorten_survey_2019}. There are several data augmentation techniques in images, namely geometric transformations such as flipping, cropping, and rotating as well as photometric transformations such as changes in brightness and contrast, color jittering, and color space conversion (HSV to grayscale) \cite{mumuni_data_2022}. In addition, there are methods that are used specifically for certain parts or regions of the image. These features allow the model to understand and focus on making the right predictions.

However, the self-supervised learning method with rotation predictions \cite{gidaris_unsupervised_2018} on images only provides insignificant increases or sometimes reduces model performance in studying many features in the data because the augmentation shifts the distribution of the data \cite{moon_tailoring_2022}. The Localizable Rotation method \cite{moon_tailoring_2022} which applies transformations to some areas of the image is able to overcome this problem. In addition, the method used in previous studies \cite{doersch_unsupervised_2015,noroozi_unsupervised_2016, gidaris_unsupervised_2018, moon_tailoring_2022} only uses one type of pretext task. This raises the question of how it affects if more than one pretext task is used and to use a gating network to unite all pretext tasks. Therefore, we propose the Gated Self-Supervised Learning method to improve image classification which consists of the localizable rotation method \cite{moon_tailoring_2022} with several additional augmentations, namely horizontal flip and randomization of RGB channels. The proposed method also uses the Mixture of Expert (MoE) architecture \cite{chen_towards_2022, yudistira_gated_2017} as a gating network in combining each augmentation task used so that the model can learn automatically and focuses more on the transformations that are most relevant and useful for supervised learning (classification).

With the Gated Self-Supervised Learning method, it is hoped that the model can learn more important features in images to help improve model performance in image classification. To test the proposed method, testing of this method will be carried out in several scenarios, namely the imbalance CIFAR dataset classification \cite{krizhevsky_learning_2009}, adversarial perturbations, classification on the Tiny-Imagenet dataset \cite{le_tiny_2015}, and semi-supervised learning.

\section{Related Works}
\label{sec:related_works}
\subsection{Self-Supervised Learning (SSL)}
\begin{figure}
    \centering
    \includegraphics{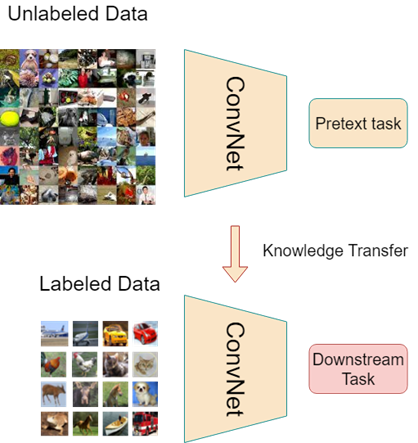}
    \caption{Self-Supervised Learning Diagram}
    \label{fig:ssl-ilustration}
\end{figure}

Self-supervised learning has received great attention in recent years, especially in the field of image recognition \cite{noroozi_unsupervised_2016, gidaris_unsupervised_2018, hendrycks_using_2019, chen_simple_2020, he_momentum_2020, lee_self-supervised_2020, zbontar_barlow_2021, moon_tailoring_2022, liu_self-supervised_2023}. This self-supervised learning method aims to help the model learn many rich features and characteristics of images by completing pre-defined tasks known as pretext tasks. After the model learns through the pretext task, the model will be fine-tuned to more specific tasks (downstream tasks) such as classification, segmentation, and object detection \cite{jing_self-supervised_2021} (see Figure \ref{fig:ssl-ilustration}).

Based on the way to determine the pretext task, the self-supervision method can be divided into two categories, namely relation-based and transformation-based tasks \cite{moon_tailoring_2022}. Relation-based works by studying features in the data to increase the similarity between positive samples and reduce the similarity for negative samples. Some examples of this type of relation-based method are MoCo \cite{he_momentum_2020}, SimCLR \cite{chen_simple_2020}, and SimSiam \cite{chen_exploring_2021}. Apart from that, the transform-based self-supervision method is also a popular method that works by creating new classes based on the data augmentation that is being performed. Some examples of transform-based methods are predicting the relative positions of image pieces \cite{doersch_unsupervised_2015}, solving jigsaw puzzles \cite{noroozi_unsupervised_2016}, predicting the degree of rotation in images \cite{gidaris_unsupervised_2018}, and localization rotation (Lorot) \cite{moon_tailoring_2022} which is a development of rotation prediction by only rotating patches in an image.

In addition, self-supervised learning methods have been applied to labeled datasets to assist supervised learning, for example SupCon \cite{khosla_supervised_2020} is a type of relation-based self-supervised learning method that utilizes labeled data where the class label indicates a positive or negative class. In addition, the self-label augmentation (SLA) \cite{lee_self-supervised_2020} method also uses labeled data to expand the label space by combining the original class label with the pseudo label that has been created. The LoRot method \cite{moon_tailoring_2022} is also a self-supervised learning method that can be applied directly to improve supervised learning performance with multi-task learning.

\subsection{Mixture of Experts (MoE)}
Mixture of Expert (MoE) is a deep learning architecture that uses a combination of more than one model that acts as a collection of experts to be able to learn complex tasks and divide them into simpler tasks in each expert model \cite{chen_exploring_2021}.
\begin{figure}
    \centering
    \includegraphics{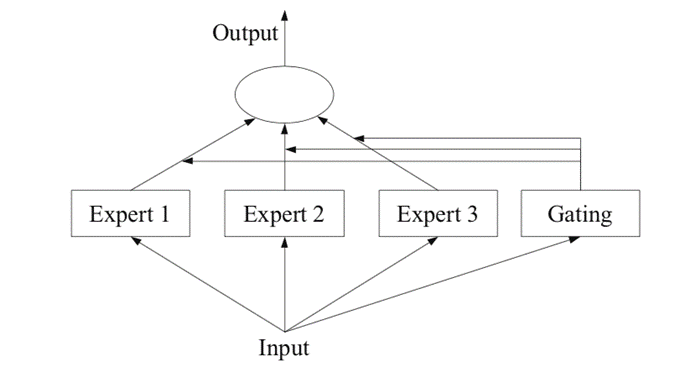}
    \caption{Mixture of Experts Diagram}
    \label{fig:moe-diagram}
\end{figure}
Figure \ref{fig:moe-diagram} shows that there are several expert models and gating networks. Mixture of Expert (MoE) has been widely applied in various studies. Sparsely-gated MoE \cite{shazeer_outrageously_2017} is one of the MoE models applied to language models and translation engines and shows large performance improvements in model capacity, training time, and model quality. In addition, MoE can be applied in video classification tasks which consist of two expert models, namely the spatial expert model and the motion (temporal) expert model \cite{yudistira_gated_2017}. Gated network is used to perform automatic weighting on each expert model (spatial and temporal) so that the weights can adaptively adjust to each expert model.

The gating network in this study was adapted from research conducted by \cite{yudistira_gated_2017} where the gating network was placed at the last layer of the backbone network to combine the loss results from each augmentation task of self-supervised learning. This method's approach is different from the previous method \cite{moon_tailoring_2022} which only uses one type of pretext task without using a gating network to weight each pretext task.

\section{Methodology}
\label{sec:methodology}
\begin{figure}
    \centering
    \includegraphics[width=\textwidth]{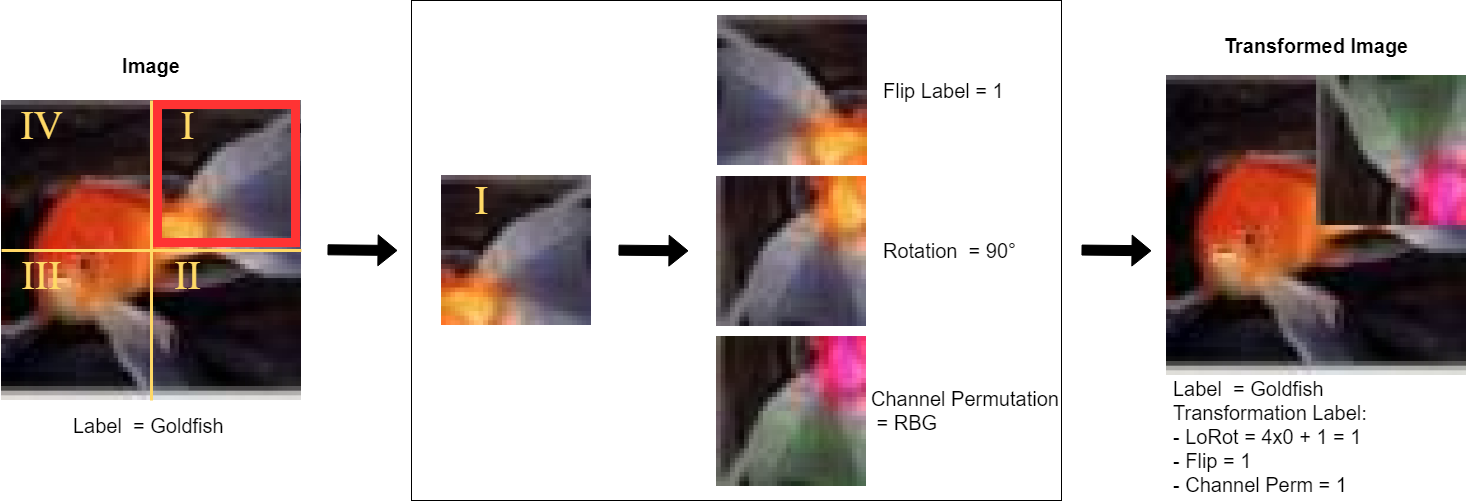}
    \caption{Transformations Diagram}
    \label{fig:transformations}
\end{figure}

The Gated Self-Supervised Learning method proposed in this study is a modification of the LoRot \cite{moon_tailoring_2022} method by using an additional transformation as a pretext task and using a gating network to combine all pretext tasks and perform automatic weighting of the pretext task. This method consists of two main parts, namely data transformation and gating network using a mixture of experts. This data augmentation and transformation is carried out so that the model can properly recognize and study the spatial features in the image. The gating network is used to help the model learn the level of importance of each transformation and augmentation used. The transformations diagram can be seen in Figure \ref{fig:transformations}.
\subsection{Transformation}
\label{subsec:transformations}
The transformation method used is useful as a pretext task by creating a pseudo label which then the model will predict the pseudo label based on the transformation used. This study uses three transformations, namely rotation, horizontal inversion, RGB channel permutation. 
\subsubsection{Rotation}
The rotation method used is the localizable rotation (LoRot-E) \cite{moon_tailoring_2022} method. This method divides the image into four quadrants and then rotates (0°, 90°, 180°, 270°) in one of the selected sections. This Transformation method produces 16 new classes which are a combination of 4 quadrants and 4 degrees of rotation.
\subsubsection{Horizontal Flip}
A horizontal flip performs a random flip over the selected quadrant of the image along the x-axis. The number of new classes generated is 2 classes.
\subsubsection{RGB Channel Permutation}
This RGB Channel Permutation is used to randomize the RGB channel in the selected image quadrant. This transformation is used to make the model study the color difference features in the image. The number of new classes generated is 6 classes (permutation 3P3)

\subsection{Gating Network}
\label{subsec:gate}
\begin{figure}[t]
    \centering
    \includegraphics{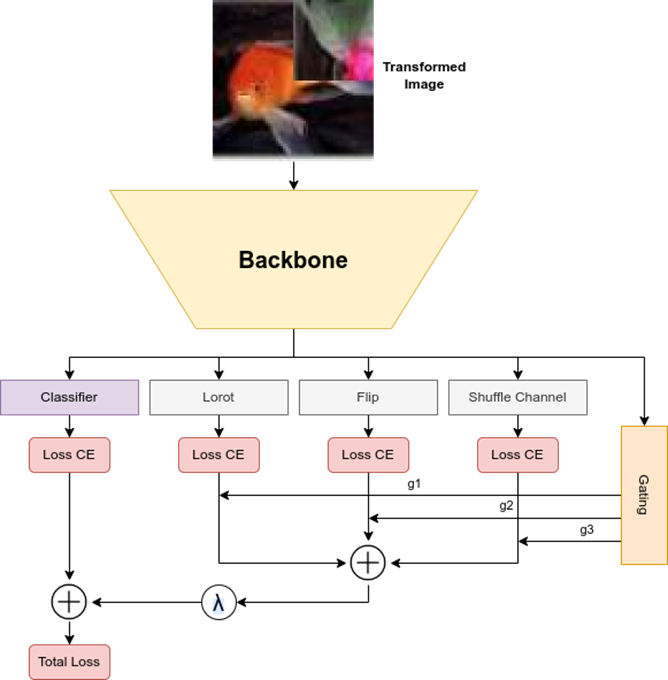}
    \caption{Gating Network Diagram}
    \label{fig:gating-network}
\end{figure}

The gate network is used to connect each pretext task and is used to study the importance of each transformation used in this research. Each transformation has a linear network as the output of the transformation class which will be weighted automatically through the gating network. It is intended that the model is able to adjust the weights automatically based on the importance of each transformation used to carry out the classification. The gating network diagram used in this study can be seen in Figure \ref{fig:gating-network}.
Each transformation has a classifier according to the output class of each transformation. The gate weight of each transformation will change dynamically using the MoE architecture adapted from \cite{yudistira_gated_2017}. The architecture used in this gating network is a fully connected layer with the softmax activation function which will be shown in equation \ref{eq:gating}.
\begin{equation}
    \label{eq:gating}
    G = softmax(W^{T}X + b) 
\end{equation}
Where, $G$ refers to the gate, $X$ is the backbone output, $W$ is the weight and $B$ is the bias. The Loss function used for all tasks, both classifier and transformation, is cross entropy loss (see equation \ref{eq:ce}).
\begin{equation}
    \label{eq:ce}
    L_{CE}=\sum_{i=0}^qy_i\log\hat{y}_i
\end{equation}
Where, $L_{CE}$ is the loss value, $y_{i}$ is the label of the image, $\hat{y}_{i}$ is the predicted class of the $q$ labels. All loss value will be combine with equation \ref{eq:loss} which is adapted from the loss function that is used in \cite{moon_tailoring_2022}.
\begin{equation}
    \label{eq:loss}
    L_{tot} = L_{C}+ \lambda \sum_{n=1}^{t}G_{n}^{T}L_{n}
\end{equation}
Where, $L_{tot}$ is the total loss value, $L_{C}$ is the classifier loss value, $L_n$ is the $n$-th SSL loss value, $G^T$ is the gate of the transformation, $t$ is the number of pretext tasks used, and $\lambda$ is the SSL ratio value. The SSL ratio is the parameter value used to set the ratio of self supervised learning (SSL) loss to classifier loss. The model will be trained by minimizing the value of the $L_{tot}$ function using the backpropagation algorithm.

\section{Experiments}
\label{sec:experiments}
We evaluate our proposed method, namely Gated Self-Supervised Learning, with various test scenarios. Tests will be carried out on cases of imbalanced CIFAR classification, adversarial perturbations, Tiny-Imagenet dataset, and semi-supervised learning scenarios. In addition, we also conducted GradCAM and T-SNE analysis on the Gated Self-Supervised Learning method. We first describe the experimental setup for this in section \ref{subsec:exp-setup} and the evaluation results of the experiments in subsequent sections. For fair comparison, we use same training configuration for all methods. In order to evaluate our proposed method, We do not only use the method with variations of rotation, horizontal flip, and channel permutation transformations, but also compare it with other transformation variations. We use rotation-horizontal flip transformation variation and rotation-channel permutation transformation variation. We also use non-gated variation (transformations only) to the experiment to see how the gating network gain an improvement to this method.
\subsection{Experimental Setup}
\label{subsec:exp-setup}
\begin{figure}[t]
    \centering
    \includegraphics[scale=0.9]{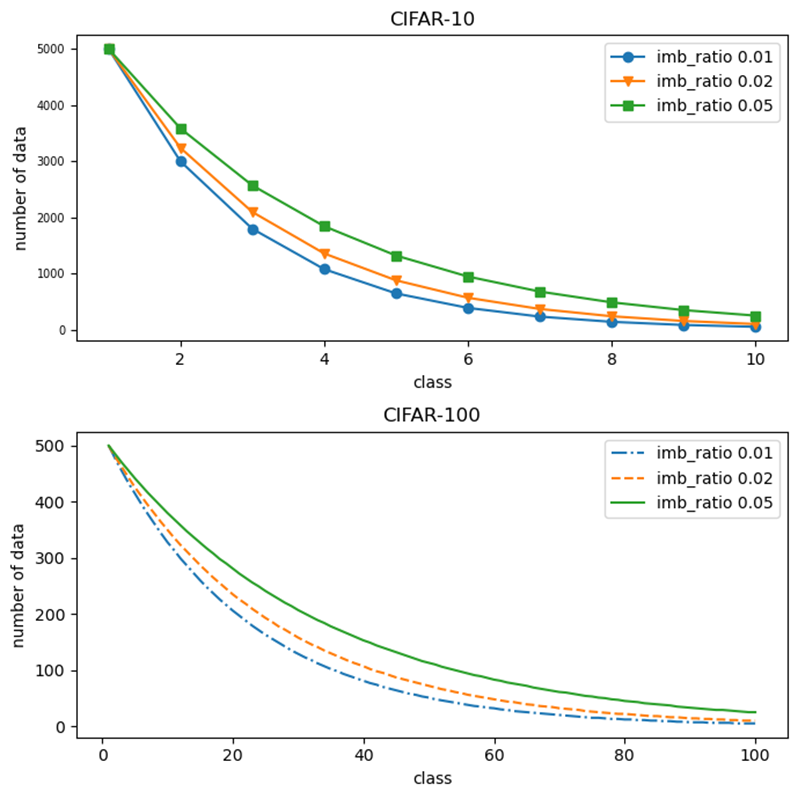}
    \caption{Distribution of Data on the CIFAR-10 and CIFAR-100 Imbalance Datasets}
    \label{fig:imb-dataset}
\end{figure}

\textbf{Imbalanced CIFAR Classification.} Tests on the imbalance CIFAR dataset classification were carried out using the LDAM-DRW (Label-Aware Margin Loss with Deferred Re-Weighting) method \cite{cao_learning_2019} as a baseline. We create an imbalance dataset scenario by modifying the amount of data for each class in the CIFAR-10 and CIFAR-100 datasets \cite{krizhevsky_learning_2009} with a certain imbalance ratio value. The value of the imbalance ratio $\rho$ is the ratio of the number of data in the fewest and most numerous classes. Then, the number of other classes is calculated by exponential decay which can be seen in equation \ref{eq:imb-dataset}.
\begin{equation}
    \label{eq:imb-dataset}
    \hat{n}_{i}=n_i\rho^{\frac{i}{K-1}}
\end{equation}
where $n_i$ and $\hat{n}_{i}$ are the number of data in the  class before and after decay and K is the number of classes in the dataset. Figure \ref{fig:imb-dataset} shows the distribution of the amount of data for each class for each imbalance ratio value used in the tests.

We use the same training configuration from \cite{cao_learning_2019} for all methods. We use ResNet-32 CIFAR varian as a backbone, 300 epochs for training, and batch size of 128. The Stochastic Gradient Descent method is used as an optimizer with a momentum of 0.9 and a weight decay of $2\times10^{-4}$. In addition, we set the initial learning rate at 0.1 which decreased by a factor of 0.01 at the 160-Th and 180-th epochs. We also set the SSL loss ratio to 0.1 for all variants from our proposed method. 

\textbf{Adversarial Perturbations.} We use PGD (Projected Gradient Descent) \cite{madry_towards_2023} training to do adversarial training in this experiment. We use the same training configurations as the following work \cite{hendrycks_using_2019}. The PGD training has several hyperparameters, namely $\epsilon$, $\alpha$ (step size), and $k$ (step). The $\epsilon$ value is $\frac{8}{255}$ for all tests. The step size $\alpha$ values used by the 10-step, 20-step, and 100-step PGD respectively are $\frac{2}{255}$, $\frac{2}{255}$, and $\frac{0.3}{255}$. The dataset used in this experiment is CIFAR-10 \cite{krizhevsky_learning_2009}. Furthermore, We use Wide Residual Network\cite{zagoruyko_wide_2016} WRN-40-2 architecture for the backbone, batch size of 128, SSL loss ratio of 0.1, SGD optimizer with momentum 0.9 and weight decay $5\times10^{-4}$, and initial learning rate 0.1 with Cosine Annealing scheduler \cite{loshchilov_sgdr_2023}. We train the model with a 10-step PGD attack and evaluate it on a 20-step and 100-step PGD attack.

\textbf{Tiny-Imagenet Classification.} For this experiments, we use same training configuration as following work \cite{avidan_automix_2022} to train the net for Tiny-Imagenet \cite{le_tiny_2015} classification. We use Residual Network \cite{he_deep_2016} 18 depth architecture  with replacing the $7\times7$ convolution and MaxPooling by a $3\times3$ convolution for the backbone, 400 training epochs, batch size of 100, and SSL loss ratio of 0.1. We also use SGD optimizer with momentum 0.9 and weight decay of 0.0001, and initial learning rate 0.2 with Cosine Annealing scheduler \cite{loshchilov_sgdr_2023}. 

\textbf{Semi-Supervised Learning.} In this experiment, we use FixMatch \cite{sohn_fixmatch_2020} as the baseline for semi supervised learning and follow the same configuration for training. We apply our proposed method, namely Gated Self-Supervised Learning, to the FixMatch which is then evaluated along with other semi-supervised learning methods. We use Wide Residual Network \cite{zagoruyko_wide_2016} WRN-28-2 architecture as the backbone, $2^{20}$ training iterations, and batch size of 64. In addition, the optimizer used is Stochastic Gradient Descent with a momentum of 0.9 and a weight decay of 0.0005. The learning rate value is set using Cosine Annealing \cite{loshchilov_sgdr_2023} with an initial learning rate value ($\eta$) of 0.03 which will decrease by $\eta \cos{\frac{7\pi k}{16K}}$ where $K$ is the total training iteration and $k$ is the current iteration. In addition, the FixMatch hyperparameter configuration used is the pseudo-label threshold $\tau$ of 0.95, unlabeled loss ratio $\lambda_u$ of 1, and $\mu$ (relative batch size of unlabeled data to labeled data) of 7. The SSL loss ratio $\lambda_{ssl}$ value used is 0.3. The semi supervised learning experiments use the CIFAR-10 \cite{krizhevsky_learning_2009} dataset with 4000 labeled data.

\textbf{GradCAM and T-SNE Analysis.} The model used in this analysis uses the Resnet-32 \cite{he_deep_2016} architecture variant of the CIFAR dataset to be trained on the CIFAR-10 dataset. Model training uses the following configuration, namely a batch size of 128, an epoch of 200, an SSL ratio of 0.1, and an initial learning rate of 0.1 which will decrease with a multiplier factor of 0.1 at the 100th and 150th epochs.

\subsection{Imbalanced CIFAR Classification}
\label{subsec:imb-classification}
\begin{table}[t]
    \centering
    \begin{tabular}{l|lll|lll}
         \hline
        Dataset                                                                         & \multicolumn{3}{l|}{CIFAR-10}                    & \multicolumn{3}{l}{CIFAR-100}                    \\ \hline
        Imbalance Ratio                                                                 & 0.01           & 0.02           & 0.05           & 0.01           & 0.02           & 0.05           \\ \hline
        LDAM-DRW                                                                        & 77.03          & 80.94          & 85.46          & 42.04          & 46.15          & 53.25          \\
        + SPP                                                                           & 77.83          & 82.13          & -              & 43.43          & 47.11          & -              \\
        + SLA-SD                                                                        & 80.24          & -              & -              & 45.53          & -              & -              \\
        + Lorot-E                                                                       & 81.82          & 84.41          & 86.67          & 46.48          & 50.05          & 54.66          \\ \hline
        \begin{tabular}[c]{@{}l@{}}+ G-SSL (LoRot-E+\\ Flip)\end{tabular}               & 81.65          & 83.93          & 86.64          & 46.48          & 49.96          & 54.63          \\
        \begin{tabular}[c]{@{}l@{}}+ G-SSL (LoRot-E+ \\ ChannelPerm)\end{tabular}       & \textbf{81.91} & 84.33          & \textbf{86.91} & \textbf{46.53} & 49.95          & \textbf{54.85} \\
        \begin{tabular}[c]{@{}l@{}}+ G-SSL (LoRot-E+\\ Flip + ChannelPerm)\end{tabular} & 81.67          & \textbf{84.65} & 86.35          & \textbf{47.37} & \textbf{50.57} & 54.75          \\ \hline
    \end{tabular}
    \caption{Classification accuracy (\%) on imbalanced CIFAR-10/CIFAR-100 dataset}
    \label{tab:imb-class}
\end{table}
We tested the Gated Self-Supervised Learning method on the CIFAR-10 and CIFAR-100 imbalance dataset cases. This test will be carried out by following the LDAM-DRW \cite{cao_learning_2019} as a baseline and its configuration. Our results will be compared with other self-supervision methods, such as SPP, SLA-SD, and LoRot-E (see Table \ref{tab:imb-class}). We also add other variants from our proposed method, such as variant Lorot-E + horizontal flip variant and Lorot-E + channel permutation variant. 

Based on Table \ref{tab:imb-class}, the Gated Self-Supervised Learning method is able to improve the performance of the model in the case of CIFAR-10 and CIFAR-100 imbalance datasets. In the CIFAR-10 dataset with an imbalance ratio of 0.05 and 0.01, the Gated Self-Supervised Learning method for the LoRot-E + channel permutation variant can improve the accuracy and performance of the model compared to the previous method, namely LoRot-E with an average of 0.165\%. In addition, in a scenario with an imbalance ratio of 0.02, the Gated Self-Supervised Learning method with all transformation variant can improve accuracy and performance by 0.24\%. In the CIFAR-100 dataset, the Gated Self-Supervised Learning method with all transformation variant is able to improve all performance and accuracy in all imbalance ratio scenarios with an average increase of 0.5\% over the LoRot-E method but the LoRot-E and channel permutation variant only able to improve accuracy and performance in the scenario of imbalance ratio of 0.01 and 0.05 with an average of 0.12\%. The LoRot-E + horizontal flip variant are not able to improve accuracy compared to the previous method, namely LoRot-E. However, all variants of the Gated Self-Supervised Learning method were able to improve the LDAM-DRW baseline with an average of 3.25\%.

\subsection{Adversarial Perturbations}
\label{subsec:adv-pert}

\begin{table}[t]
    \centering
    \begin{tabular}{l|lll}
        \hline
        \multirow{2}{*}{Method}                & \multicolumn{3}{l}{Dataset} \\ \cline{2-4} 
                                               & clean  & 20-step & 100-step \\ \hline
        Normal Training                        & 95.3   & 0       & 0        \\ \hline
        PGD Training                           & 83.4   & 46.5    & 46.5     \\
        + rotations                            & 82.8   & 49.3    & 49.2     \\
        + LoRot-E                              & 82.6   & 52.8    & 52.8     \\ \hline
        + LoRot-E + Flip + ChannelPerm         & 82.3   & 52.4    & 52.4     \\
        + G-SSL (LoRot-E + Flip)               & 81.0   & 52.5    & 52.4     \\
        + G-SSL (LoRot-E + ChannelPerm)        & 81.94  & 52.6    & 52.6     \\
        + G-SSL (LoRot-E + Flip + ChannelPerm) & 81.7   & \textbf{53.2}    & \textbf{53.3 }    \\ \hline
    \end{tabular}
    \caption{Classification accuracy (\%) of the adversarial attack on the CIFAR-10 dataset}
    \label{tab:adv_pert}
\end{table}
Adversarial perturbations are one of the weaknesses of deep neural networks where a slight change in the input causes the network to misclassify \cite{madry_towards_2023}. In this experiment, We test the Gated Self-Supervised Learning method on adversarial attack cases to test the model's performance and robustness. We use PGD training \cite{madry_towards_2023} as a baseline and use the same training configuration as \cite{hendrycks_using_2019}.  Our results will be compared with the normal training, baseline, and other self-supervision methods, such as rotations \cite{hendrycks_using_2019} and LoRot-E \cite{moon_tailoring_2022}. We also include other variants from our proposed method, such as variant Lorot-E + horizontal flip variant and Lorot-E + channel permutation variant and non-gating network variant.

Table \ref{tab:adv_pert} 6.2 shows that the Gated Self-Supervised Learning method is able to improve the performance of the model in the case of an adversarial attack even though there is a decrease in accuracy in the clean dataset. This can be seen from the increased accuracy of the Gated Self-Supervised Learning method for all transformation variants with an average of 0.45\% against the LoRot-E method for 20-step and 100-step PGD. Although the other Gated Self-Supervised Learning variants have not been able to improve performance compared to the LoRot-E method, all variants have been able to improve model performance at  PGD training baseline with an average of 6.26\%. In addition, the use of the gating network can improve model performance which can be seen in the combined variant of all transformations without the gating network causing a decrease in accuracy compared to the LoRot-E method itself.

\subsection{Tiny-Imagenet Classification}
\label{subsec:tiny-imagenet}

\begin{table}[t]
    \centering
    \begin{tabular}{l|l}
        \hline
        Method                                & Validation Accuracy \\ \hline
        baseline                              & 61.08                                    \\ \hline
        +MixUp                                & 63.86                                    \\
        +CutMix                               & 65.53                                    \\
        +SmoothMix                            & 66.65                                    \\
        +GridMix                              & 65.14                                    \\ \hline
        +LoRot-E                              & 66.52                                    \\ \hline
        +LoRot-E + Flip + ChannelPerm         & 66.23                                    \\
        +G-SSL (LoRot-E + Flip)               & 66.34                                    \\
        +G-SSL (LoRot-E + ChannelPerm)        & \textbf{67.27}                           \\
        +G-SSL (LoRot-E + Flip + ChannelPerm) & \textbf{67.41}                           \\ \hline
    \end{tabular}
    \caption{Classification accuracy (\%) on the Tiny-Imagenet dataset}
    \label{tab:tiny-imagenet}
\end{table}

Furthermore, we test this Gated Self-Supervised Learning method on the Tiny-Imagenet dataset \cite{le_tiny_2015}. It aims to test the ability of this method on a subset of the Imagenet dataset. Then, we compare our results with LoRot-E \cite{moon_tailoring_2022} and the MixUp \cite{zhang_mixup_2018} augmentation method with other variants, such as CutMix \cite{yun_cutmix_2019}, SmoothMix \cite{lee_smoothmix_2020}, and GridMix \cite{baek_gridmix_2021}. We also include other variants from our proposed method, such as variant Lorot-E + horizontal flip variant and Lorot-E + channel permutation variant along with the non-gating network variant.

Based on Table \ref{tab:tiny-imagenet}, the results of testing the Gated Self-Supervised Learning method on the Tiny-Imagenet dataset are able to improve the performance of the model compared to other methods. This is indicated by an increase in accuracy in the Gated Self-Supervised Learning method for all transformation variants and Lorot-E + channel permutation variant with an average of 0.82\% compared to the LoRot-E method. Although the flip variant of the Gated Self-Supervised Learning method has less accuracy than the LoRot-E method, all variants of the Gated Self-Supervised Learning method are able to increase the accuracy of the baseline Resnet-18 by 5.93\%. The Gated Self-Supervised Learning all transformations variant and Lorot-E + channel permutation variant is also able to outperform the MixUp \cite{zhang_mixup_2018} and its variants, such as CutMix \cite{yun_cutmix_2019}, SmoothMix \cite{lee_smoothmix_2020}, and GridMix \cite{baek_gridmix_2021}. 

\begin{figure}
    \centering
    \includegraphics[width=\textwidth]{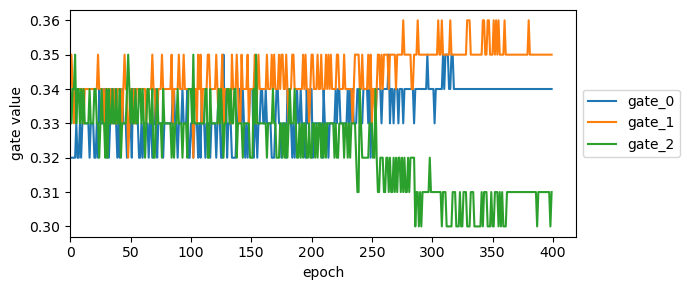}
    \caption{Gating network values for each transformation loss}
    \label{fig:gate-value}
\end{figure}

In addition, based on the test results it is shown that the gating network is able to improve model performance which can be seen in the variant combination transformation without gating network reducing the performance and accuracy of the model compared to the LoRot-E method. Therefore, it is necessary to plot the gate value of the gating network in training. This is to see how each gate value from each transformation (pretext task) adapts automatically to each training epoch. The graph of the values of the gating network can be seen in Figure \ref{fig:gate-value}. Gate\_0, gate\_1, and gate\_2 respectively are the gating values of LoRot-E, horizontal flip, channel permutation of RGB. During training prior to epoch 250, gate values ranged from 0.32 to 0.3. After epoch 250, the gating values ranged from 0.3 to 0.36.

\subsection{Semi-Supervised Learning}
\label{subsec:semi-sl}

In addition to the classification cases that have been tested in the previous subsection, we test this Gated Self-Supervised Learning method in a semi-supervised learning scenario. Semi-supervised learning is a new paradigm for building models by utilizing both labeled data and unlabeled data to enhance supervised learning due to the limitations of labeled data. This experiment is conducted to find out how the performance of our proposed method is in handling both labeled and unlabeled data. The Gated Self-Supervised Learning method is applied in one of the semi-self-supervised learning methods, namely FixMatch \cite{sohn_fixmatch_2020} to improve the performance of the model. We make modifications by adding SSL loss to the total loss in the FixMatch method (see equation \ref{eq:mod-fixmatch}).
\begin{equation}
    \label{eq:mod-fixmatch}
    L_{tot}=L_s+\lambda_uL_u+\lambda_{ssl}L_{ssl}
\end{equation}
where $L_{tot}$ is the total loss, $L_s$ is the loss from labeled data classification, $L_u$ is the unlabeled loss of unlabeled data from the FixMatch method, $L_{sal}$ is the SSL loss, along with $\lambda_u$ and $\lambda_{sal}$ are the ratio values of the unlabeled loss and SSL loss. The $L_ssl$ value is calculated using the Gated Self-Supervised Learning method which is applied to all data, both labeled and unlabeled.

\begin{table}[t]
    \centering
    \begin{tabular}{l|l}
        \hline
        Method           & Accuracy \\ \hline
        $\Pi$-Model          & 85.99    \\
        Pseudo-Labelling & 83.91    \\
        Mean Teacher     & 90.81    \\
        MixMatch         & 93.58    \\
        FixMatch         & 95.74    \\ \hline
        FixMatch+G-SSL   & 94.95    \\ \hline
    \end{tabular}
    \caption{Classification accuracy on semi-supervised learning with the 4000 labeled CIFAR-10 dataset}
    \label{tab:semi-sl}
\end{table}

Based on the results in Table \ref{tab:semi-sl}, the use of the Gated Self-Supervised Learning method cannot improve from the previous baseline method, namely FixMatch \cite{sohn_fixmatch_2020}. This is shown because of a decrease in the accuracy value of 0.79\%. Nevertheless, the Gated Self-Supervised Learning method still shows improvement compared to other methods such as $\Pi$-Model \cite{rasmus_semi-supervised_2015}, Pseudo-Labeling \cite{lee_pseudo-label_2013}, Mean Teacher \cite{tarvainen_mean_2017}, and MixMatch \cite{berthelot_mixmatch_2019}. The average increase in accuracy is 6.38\%.

\subsection{GradCAM and T-SNE Analysis}
\label{subsec:gradcam-tsne}
We conduct an analysis of the Gated Self-Supervised Learning method using Grad-CAM \cite{selvaraju_grad-cam_2020} and T-SNE \cite{maaten_visualizing_2008}. Grad-CAM is used to find out how the model performs classification by using gradients in the last layer to find the most important parts or image features that influence classification. T-SNE is used to visualize data with high dimensions which will be projected into smaller dimensions. This method works on the model by reducing the output dimensions of the embedding representation layer before the classification layer is visualized. The Gated Self-Supervised Learning method will be compared with the LoRot-E method, Vanilla (normal training), and transformation variations without a gating network. The Gated Self-Supervised Learning variant used is all transformations (LoRot-E, horizontal inversion, and RGB channel permutation). For Grad-CAM, the results of the activation map from each method will be compared. For T-SNE, the dimensional reduction results of each method will be visualized and compared.

\begin{figure}[t]
    \centering
    \includegraphics[scale=0.5]{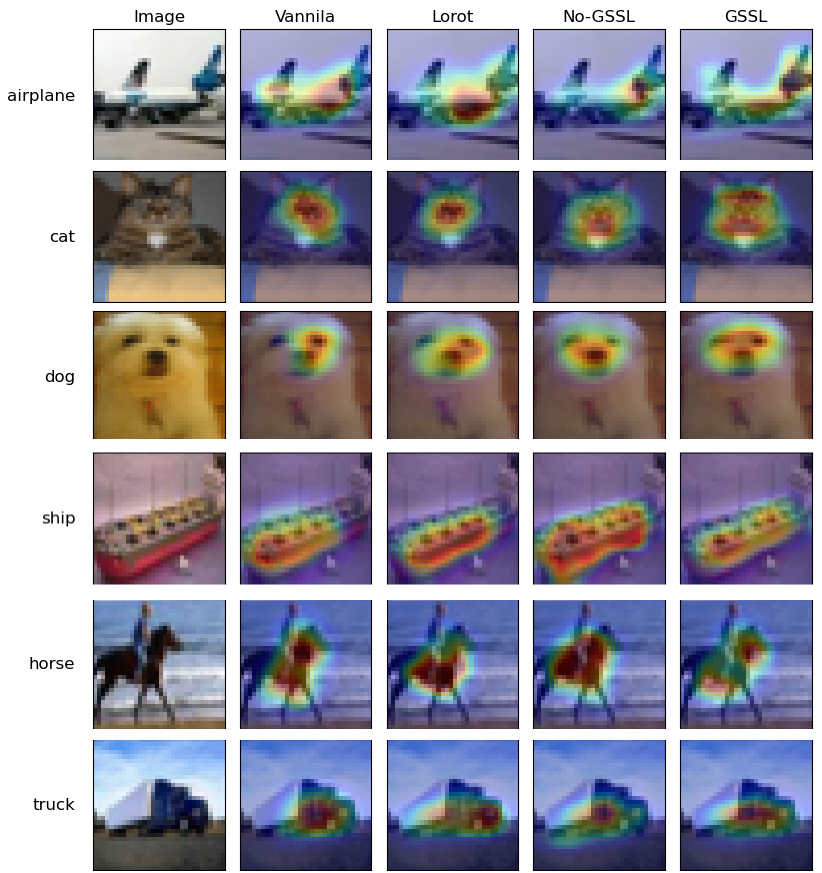}
    \caption{GradCAM activation map results from vanilla, Lorot-E, Gated Self-Supervised Learning, and non-gating variant}
    \label{fig:gradcam}
\end{figure}
We apply Grad-CAM to the Gated Self-Supervised Learning Grad-CAM method on the last convolution layer. This is used to find out how the model generates one prediction from one image to determine the parts or features in the image that are most important and have a major influence in classifying. The result of Grad-CAM is an activation map that will be overlay on the original image. The results of the activation map from Grad-CAM can be seen in Figure  below. Based on Figure \ref{fig:gradcam} it can be seen that the model trained with the Gated Self-Supervised Learning method is able to pay attention and focus on the features or parts of the image that are important in classification. For example, in the labeled images of cats and dogs, the model trained with the Gated Self-Supervised Learning method focused more on the face in the image than the other methods only on some parts of the face. In addition, in the image of ship and aircraft labels, the Gated Self-Supervised Learning method is able to recognize and focus on objects more broadly and thoroughly compared to other methods which only partially or exceed objects.

\begin{figure}[t]
    \centering
    \includegraphics[scale=0.8]{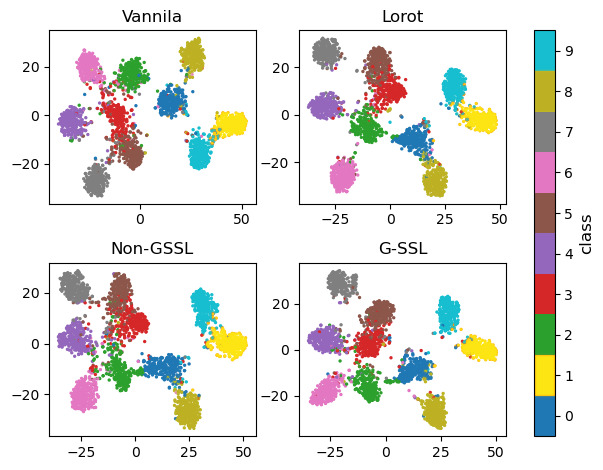}
    \caption{Visualization of T-SNE results from vanilla, LoRot-E, Gated Self-Supervised Learning, and non-gating variant}
    \label{fig:tsne}
\end{figure}
We also perform T-SNE analysis to see how the results of the layer embedding representation are based on the Gated Self-Supervised Learning method and other methods. In addition, this analysis was carried out to see how each class in the image can be represented by an embedding layer where data with each of the same class will be more collected in a group. Figure \ref{fig:tsne} shows the scatter plot diagram of the T-SNE visualization results of each of the compared methods, namely plain training (vanilla), LoRot-E, variations of all transformations without gating network (LoRot-E, horizontal reversal, and channel permutations RGB), and Gated Self-Supervised Learning (LoRot-E, horizontal inversion, and permutation of RGB channels). Based on Figure \ref{fig:tsne}, the Gated Self-Supervised Learning method is able to group data with the same class better than other methods. This can be seen in class 9 and 1 which overlap with other methods but Gated Self-Supervised Learning is able to separate them. In addition, classes 2, 3, with 5, and 8 with 0 are also able to be separated by the Gated Self-Supervised Learning method better than the other methods. The Gated Self-Supervised Learning method is also better at grouping data that has the same class. This can be seen in classes 3, 5, and 0 which are more spread out in other methods.

\section{Conclusions}
\label{sec:conclusions}
The method proposed in this study is the Gated Self-Supervised Learning method which is used to improve the image classification task. This method is the development of the Self-Supervised Learning method by using more than one transformation (LoRot-E, horizontal reversal, and RGB channel permutation) as a pretext task and using a gating network to combine the loss values of each transformation. The results show that the Gated Self-Supervised Learning method is able to improve performance in image classification tasks in various scenarios, namely the CIFAR imbalance dataset, adversarial perturbation, and the Tiny-Imagenet dataset although there is a decrease in performance in the case of semi-supervised learning, especially in the FixMatch method but still has better performance than other methods. In addition, in the analysis of Grad-CAM and T-SNE, the Gated Self-Supervised Learning method is able to identify important features and parts of the image that influence image classification and is able to represent data for each class and separate different classes properly.

There are several suggestions for further research. Subsequent research can test the Self-Supervised Learning method by adding new transformations to multiply different pretext tasks. In addition, this Self-Supervised Learning method needs to be tested on large datasets such as ImageNet to test the performance of the method on large datasets. This method is applied to other tasks and datasets such as the health, agriculture, and other fields related to image classification.



	
\def\UrlBreaks{\do\/\do-}

 \bibliographystyle{elsarticle-num}
 \bibliography{cas-refs}





\end{document}